\documentclass[sigconf]{acmart}

\acmConference[SemRec '21]{SemRec '21}{2021}{ISWC}
\acmBooktitle{}
\acmPrice{15.00}
\acmISBN{978-1-4503-XXXX-X/18/06}

\begin{document}

\title{Why Settle for Just One? \\ Extending $\mathcal{E}\mathcal{L}^{++}$ Ontology Embeddings with Many-to-Many Relationships}

\author{Biswesh Mohapatra}
\affiliation{
  \institution{International Institute of Information Technology}
  \country{Bangalore, India}}
\email{biswesh.mohapatra@iiitb.org}

\author{Sumit Bhatia}
\affiliation{
  \institution{Adobe Research}
  \country{India}}
\email{sumit.summit@gmail.com}

\author{Raghava Mutharaju}
\affiliation{
 \institution{Indraprastha Institute of Information Technology}
 \country{Delhi, India}}
 \email{raghava.mutharaju@iiitd.ac.in}

\author{G. Srinivasaraghavan}
\affiliation{%
  \institution{International Institute of Information Technology}
  \country{Bangalore, India}
  }
\email{gsr@iiitb.ac.in}  

\renewcommand{\shortauthors}{Mohapatra, et al.}

\begin{abstract}
  Knowledge Graph (KG) embeddings provide a low-dimensional representation of entities and relations of a Knowledge Graph and are  used successfully for various applications such as question answering and search, reasoning, inference, and missing link prediction. However, most of the existing KG embeddings only consider the network structure of the graph and ignore the semantics and the characteristics of the underlying ontology that provides crucial information about relationships between entities in the KG. Recent efforts in this direction involve learning embeddings for a Description Logic (logical underpinning for ontologies) named $\mathcal{EL}^{++}$. However, such methods consider all the relations defined in the ontology to be one-to-one which severly limits their performance and applications. We provide a simple and effective solution to overcome this shortcoming that allows such methods to consider many-to-many relationships while learning embedding representations. Experiments conducted using three different $\mathcal{EL}^{++}$ ontologies show substantial performance improvement over five baselines.  Our proposed solution also paves the way for learning embedding representations for even more expressive description logics such as $\mathcal{SROIQ}$. 
\end{abstract}



\keywords{NeuroSymbolic AI, Ontology, Embeddings}


\maketitle

\section{Introduction}
The availability of data along with the computational power has led to the rise of deep learning which has become an important tool in various fields like computer vision and natural language processing. Despite the huge success, artificial neural networks have not been able to perform well on reasoning and logical tasks. They suffer from issues such as explainability and robustness. Knowledge Graphs~\cite{hogan2021knowledge} and ontologies~\cite{Guarino} are  structural knowledge bases that capture the relationship between different entities making them suitable for reasoning tasks. Unfortunately they are not ideal for real time usage as reasoning time, especially on complex and large ontologies could have worst-case exponential time complexity. Hence there is a trade-off between expressivity and complexity of such reasoners. 

Knowledge Graph (KG) embeddings are an effort to combine the knowledge present in the Knowledge Graphs with the generalisation capability of the neural networks~\cite{Singh_Mondal_Bhatia_Mutharaju_2021}. KG embeddings learn embedding functions that map the entities of Knowledge Graphs to vector space. Different Learning methods for such embeddings have been proposed~\cite{transh,transe,yang2015embedding,Nickel@icml,pmlr-v48-trouillon16,Lin2015} that try to preserve various properties of these knowledge bases. In case of ontologies, they help create approximate reasoners which could reason over complex ontologies while having low time complexity.

Description logics~\cite{description_primer} are a family of knowledge representation languages that provide the formal underpinning for Web Ontology Language (OWL). OWL 2, which is the most recent version of OWL, is used to build ontologies. They provide the schema information for the knowledge captured in KGs. 

Most of the KG embeddings fail to consider the underlying constraints and characteristics of the ontologies. Hence, reasoning tasks do not perform well on such embeddings of ontologies. In order to tackle this issue, Kulmanov et al.~\cite{Kulmanov} proposed EL Embeddings (EmEl) that incorporate the geometric structure of $\mathcal{E}\mathcal{L}^{++}$ description logic ontologies into the embeddings. Mondal et al. \cite{Mondal} later added role oriented $\mathcal{E}\mathcal{L}^{++}$ constructs into the embeddings through their proposed EmEl$^{++}$ method. While these methods have provided a new technique to perform reasoning tasks on the ontologies, they have a fundamental issue that restricts their performance on $\mathcal{E}\mathcal{L}^{++}$ ontologies and restricts them from being used in more complex description logic based ontologies such as $\mathcal{SROIQ}$, which is the basis for OWL 2 DL, a fragment of OWL 2. We provide a simple and effective way to convert the embedding functions such that the roles (relation equivalent in ontologies) can be considered as many-to-many instead of one-to-one functions as is in the case of EmEl. This is significant as most of the roles in ontologies connect a class to multiple classes. For example, the \emph{fatherOf} role can connect an individual to multiple individuals if he is the father of all of them. This issue becomes more important when we try to move to complex description logics like $\mathcal{SROIQ}$ which has properties such as cardinality that depend on such many-to-many roles. 

\begin{figure*}[ht]
\centering
\includegraphics[scale=0.55]{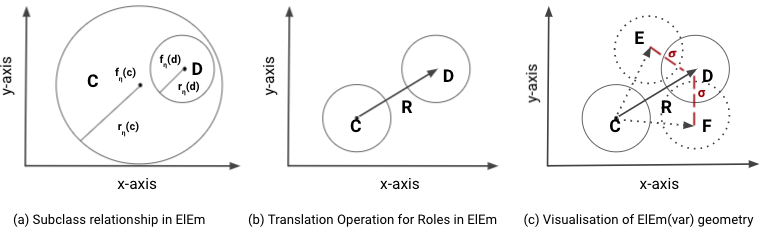}
\vspace{-2ex}
\caption{\label{fig:geometric} Geometric representation of classes and relations. (a) shows the representation of subclass relation where D $\sqsubseteq$ C and thus the n-ball of D lies inside n-ball of C. (b) shows class C getting translated to D using relation R for a tuple (C,R,D) in the ontology. (c) In case of EmEl(var), the variance $\sigma$ lets the the entity C relate to multiple entities with the relation R. Any entity which falls within $\sigma$ distance of C+R are also related to C through R. E and F are the boundary entities for C and R.} 
\end{figure*}

\noindent {\bf Contributions: } (1) We provide a simple method to incorporate many-to-many roles in tranlation based embeddings like TransE~\cite{transe}. (2) We show how the method could be used to modify the ontological embedding EmEl~\cite{Kulmanov}. (3) We demonstrate the effectiveness of the method on three publicly available ontologies. (4) Our work provides a foundation for work on complex DL reasoners. 

\section{Related Work}
Node2Vec~\cite{node2vec} used the concept of representing facts as triples of the form (h,r,t) which became a standard for various other models. Similarity based scoring functions were used in this work. In order to capture the underlying properties of the knowledge bases TransE~\cite{transe} considered the relations in a KG as a translation operator over the entities and used distance based scoring functions. TransH~\cite{transh} further allowed the relations to be many-to-many and reflexive by modelling the relations as a hyperplane in the vector space. DistMult~\cite{yang2015embedding} on the other hand used matrix factorisation to relate various entities.  Existing works on ontology embedding such as Onto2vec~\cite{ontovec} focuses on using word2vec as an underlying model. While the work focuses on encoding the entities and relations, it is unable to handle complex relations in an ontology. \cite{Ebrahimi} provide neuro-symbolic deep deductive reasoners for $\mathcal{E}\mathcal{L}^{++}$ DL and first-order logic. \cite{ijcai2020-252} pointed out that geometric models are a better way to learn embeddings for ontologies. The simplicity of the translation based models for KG embeddings~\cite{transe,transh,yang2015embedding} to measure the correctness of a fact as a distance between entities after being translated by the relation made them popular. EmEl~\cite{Kulmanov} and EmEl$^{++}$~\cite{Mondal} used this translation technique to create embeddings for ontologies which preserve their underlying structures and characteristics. In order to accomplish it, the models use geometric models to learn embeddings. The classes are considered to be n-balls in an n-dimensional space which are translated by the relation vectors to the n-ball of the corresponding class of the fact. This geometric structure provides a way to incorporate various structural properties of an ontology eg. subclass properties. However, like TransE, these models too restrict their triplets to a one-to-one mapping. Not only do these restrictions affect the performance of these models on $\mathcal{E}\mathcal{L}^{++}$ ontologies but also restrict them from being used in more complex description logics such as $\mathcal{SROIQ}$.

\section{Background on Ontology Embeddings}
\label{sec:emel}
Kulmanov et al.~\cite{Kulmanov} introduced the concept of incorporating geometric structure of ontologies into the embeddings. They proposed embeddings for the $\mathcal{E}\mathcal{L}^{++}$ description logic (EmEL) that captures the underlying structures and characteristics of the ontology by treating ontology classes as n-balls in n-dimensional space. These n-balls are represented by a center which is a n-dimensional vector and a radius which is a scalar. The relations in the ontology are considered as n-dimensional vectors which are used to translate the class from one point in the space to another. The center and the radius of each class (n-ball), along with the relations can be learnt over multiple iterations. They make up the embeddings for the ontology. Figure~\ref{fig:geometric}(a) and Figure~\ref{fig:geometric}(b) show the geometric representation of classes and relations in 2-dimensional space.

Hence, they define a geometric ontology embedding $\eta$ as a pair ($f_\eta$, $r_\eta$) of functions that map classes and relations in ontology O into $R^n$. 
Thus $f_\eta$ : C $\cup$ R  $\mapsto \mathbb{R}^n$ and $r_\eta$ : C $\mapsto \mathbb{R}$. Here C is a class and R is a relation and O is defined as ($\mathbb{C}, \mathbb{R'}, \mathbb{I}$; ax) where $\mathbb{I}$ are individual symbols, $\mathbb{C}$ is set of class symbols, $\mathbb{R'}$ is set of relation symbols and ax are axioms (facts). Basically, $f_\eta$(c) represents center of C, $r_\eta$(c) represents radius of C and $f_\eta$(r) represents vector of R.

Each axiom, ax, is transformed into its equivalent normal form using a set of conversion rules from~\cite{baader}. These rules help transform the set of axioms in the ontology into one of four forms without any loss of information. These are
(1) Subclass axiom: $C \sqsubseteq D$ (2) Intersection axiom: $C \sqcap D \sqsubseteq E$ (3) Existential restriction (right-hand side): $C \sqsubseteq \exists R. D$ and (4) Existential restriction (left-hand side): $\exists R.C \sqsubseteq D$ \noindent where C, D, E $\in \mathbb{C}$ and R $\in \mathbb{R}$.

EmEl formulates a loss function for each of the four normal forms in order to preserve the semantics of $\mathcal{EL}^{++}$ in the embeddings. The loss functions are as follows.

\begin{equation}
    \begin{aligned}
    \text{loss}_{C \sqsubseteq D} (c, d) = {} & \text{max}(0, \parallel f_\eta(c) - f_\eta(d) \parallel + r_\eta(c) - r_\eta(d) - \gamma ) \\
    & + | \parallel f_\eta(c) \parallel - 1 | + | \parallel f_\eta(d) \parallel - 1 |
    \end{aligned}
    \label{eqn:1}
\end{equation}

In Eqn \ref{eqn:1}, we try to preserve the subclass property of the entities. Here the euclidean distance between the centers of C and D should be less than the difference between the radius of D and C. Once this is achieved, we ensure that the n-ball representing D is bigger than that of C and that n-ball of C lies completely inside D. Here $\gamma$ is a hyperparameter called margin.  $| \parallel f_\eta(c) \parallel - 1 | + | \parallel f_\eta(d) \parallel - 1 |$ ensures that the n-balls lie in the unity sphere.

\begin{equation}
    \begin{aligned}
    \text{loss}_{C \sqcap D \sqsubseteq E}  (c, d, e) = {} & \text{max}(0, \parallel f_\eta(c) - f_\eta(d) \parallel - r_\eta(c) - r_\eta(d) - \gamma ) \\
    & + \text{max}(0, \parallel f_\eta(c) - f_\eta(e) \parallel - r_\eta(c) - \gamma ) \\
    & + \text{max}(0, \parallel f_\eta(d) - f_\eta(e) \parallel - r_\eta(d) - \gamma ) \\
    & + | \parallel f_\eta(c) \parallel - 1 | + | \parallel f_\eta(d) \parallel - 1 | \parallel f_\eta(e) \parallel - 1 | 
    \end{aligned}
    \label{eqn:2}
\end{equation}

In Eqn \ref{eqn:2}, we incorporate the intersection property. The first term ensures that C and D are not disjoint sets. While second and third terms force the center of E to lie in the intersection of D.

\begin{equation}
    \begin{aligned}
    \text{loss}_{C \sqsubseteq \exists R. D} (c, d, r) = {} & \text{max}(0, \parallel f_\eta(c) + f_\eta(r) - f_\eta(d) \parallel \\ 
             &  \qquad \quad + r_\eta(c) - r_\eta(d) - \gamma ) \\
    & + | \parallel f_\eta(c) \parallel - 1 | + | \parallel f_\eta(d) \parallel - 1 |
     \end{aligned}
     \label{eqn:3}
\end{equation}

\begin{equation}
    \begin{aligned}
    \text{loss}_{\exists R.C \sqsubseteq D} (c, d, r) = {} & \text{max}(0, \parallel f_\eta(c) - f_\eta(r) - f_\eta(d) \parallel \\
     & \qquad \quad - r_\eta(c) - r_\eta(d) - \gamma ) \\
    & + | \parallel f_\eta(c) \parallel - 1 | + | \parallel f_\eta(d) \parallel - 1 |
     \end{aligned}
     \label{eqn:4}
\end{equation}

Eqn \ref{eqn:3} and Eqn \ref{eqn:4} describe the loss function for the third and fourth normal forms respectively. Every point that lies within an n-ball representing a class is a potential instance of that class. The loss functions capture this by applying relations as translations on these points (following the TransE~\cite{transe} relation model). The relation vector $f_\eta(r)$ when added to the center of class C should be at a maximum distance of the sum of radii of C and D from the center of D. Eqn \ref{eqn:4} reverses the direction of translation from Eqn \ref{eqn:3}. 
 
\begin{equation}
    \begin{aligned}
    \text{loss}_{C \sqcap D \sqsubseteq \perp} (c, d) = {} & \text{max}(0, r_\eta(c) + r_\eta(d) - \parallel f_\eta(c) - f_\eta(d)  \parallel + \gamma ) \\
    & + | \parallel f_\eta(c) \parallel - 1 | + | \parallel f_\eta(d) \parallel - 1 |
     \end{aligned}
     \label{eqn:5}
\end{equation}

Eqn \ref{eqn:5} describes the loss function for disjoint classes C and D while Eqn \ref{eqn:6} refers to the specific loss function for bottom class whose radius must be equal to zero.

\begin{equation}
    \text{loss}_{C \sqsubseteq \perp} (c) = r_\eta(c)
    \label{eqn:6}
\end{equation}

EmEl++~\cite{Mondal} added role constructors to EmEl. The translation approach of EmEl and EmEl++ is not suitable for more expressive description logics such as $SROIQ$ because the translation operator on relations makes them one-to-one. This limits the capabilities of the model as most  of the relations are many-to-many. We propose a modification to their approach, named  EmEl(var), to overcome the issue.

\section{Proposed Approach}
In order to address the one-to-one relation restriction, we used a simple yet powerful technique that provides a foundation for further work in embeddings based description logic reasoning. We consider the relations to have a variance (uncertainty) leading to the translation having various possible regions in the vector space. This lets us model one-to-many and many-to-many relations in the ontology. As a result, we can model complex properties such as cardinality.

We consider the variance to be a hard bound. The translation of n-ball C on relation vector R could now be within $\sigma$ distance of n-ball D in a tuple (C, R, D) where C, D $\in$ $\mathbb{C}$ and R $\in$ $\mathbb{R'}$. Hence all the points within $\sigma$ distance from the translated space are related to C through R. This removes the one-to-one limitation of previous methods. We call this model EmEl(var).


Every relation has its own $\sigma$ which is learnt during training.  In order to avoid $\sigma$ from becoming infinite, we keep the absolute value of $\sigma$ as a loss component for regularisation. Figure \ref{fig:geometric}(c) shows a visual representation of EmEl(var).

Hence, the definition of the geometric ontology embedding $\eta$ now becomes a tuple ($f_\eta$, $r_\eta$, $\sigma_\eta$) of functions that map classes and relations in ontology O into $R^n$, where $f_\eta$ : C $\cup$ R  $\mapsto \mathbb{R}^n$, $r_\eta$ : C $\mapsto \mathbb{R}$ and $\sigma_\eta$ : R $\mapsto \mathbb{R}$. The modified loss functions are provided in Eqn \ref{eqn:5_3} and Eqn \ref{eqn:5_4}. Note that the loss function for other normal forms remain the same as in EmEl++.

\begin{equation}
    \begin{aligned}
    \text{loss}_{C \sqsubseteq \exists R. D} (c, d, r) = {} & \text{max}(0, \parallel f_\eta(c) + f_\eta(r) - f_\eta(d) \parallel \\
    & \qquad \quad + r_\eta(c) - r_\eta(d) - \sigma_\eta(r) - \gamma ) \\
    & + | \parallel f_\eta(c) \parallel - 1 | + | \parallel f_\eta(d) \parallel - 1 | + \sigma_\eta(r)
     \end{aligned}
     \label{eqn:5_3}
\end{equation}
 \\
\begin{equation}
    \begin{aligned}
    \text{loss}_{\exists R.C \sqsubseteq D} (c, d, r) = {} & \text{max}(0, \parallel f_\eta(c) - f_\eta(r) - f_\eta(d) \parallel \\
    & \qquad \quad - r_\eta(c) - r_\eta(d) - \sigma_\eta(r) - \gamma ) \\
    & + | \parallel f_\eta(c) \parallel - 1 | + | \parallel f_\eta(d) \parallel - 1 | + \sigma_\eta(r)
     \end{aligned}
     \label{eqn:5_4}
\end{equation}


\section{Experiments}
\noindent \textbf{Datasets:} We use the follwoing publicly available ontologies of varying sizes.
\begin{enumerate}
    \item GALEN~\cite{galen} captures clinical information. It consists of 84,537 axioms with 1,010 relations and 24,353 classes.
    \item Gene Ontology (GO)~\cite{goc} has a unified representation of genes across all species. It consists of 130,094 axioms with 45,907 classes and 16 relations.
    \item SNOMED CT~\cite{snomed} is a comprehensive ontology of clinical terms with 989,186 axioms, 307,712 classes and 60 relations.
\end{enumerate}

\noindent \textbf{Baselines:}
\begin{enumerate}
    \item TransE~\cite{transe}. This model introduced the idea of translation into knowledge graph embeddings where relations help translate the entity vectors to another related entity in the vector space.
    \item TransH~\cite{transh}. In order to overcome the one-to-one problem as seen in TransE, it considers relations as hyperplanes and uses projection of entities on those hyperplanes to relate to other entities. 
    \item DistMult~\cite{yang2015embedding}. A matrix factorization based model which has empirically performed well on compositional reasoning tasks.
    \item EmEl~\cite{Kulmanov}. The embeddings take into consideration the structure of ontologies.
    \item EmEl++~\cite{Mondal}. An incremental work on top of EmEl with added properties like role inclusion and role chain. 
\end{enumerate}

\subsection{Model Training}
In order to train the embeddings, we use Pytorch~\cite{pytorch} and its embedding layers. 
Pykeen framework~\cite{ali2020keen} was used for implementing TransE, TransH, and DistMult embedding models. For EmEl embeddings, source code provided by the authors was used. In order to learn the embeddings for different models, we first normalize the ontologies, i.e., convert the axioms into one of the four normal forms discussed in Section~\ref{sec:emel}. 
All individuals in the ontology are considered as nominal classes (containing one instance) and the embeddings learn to make their radii zero resulting in a point in the vector space.  Next, we remove some of the subclass relation pairs for validation (20\%) and testing (10\%). Remaining 70\% sub-class relation pairs are used for training the embedding functions. 

\subsection{Evaluation Metrics}

Subsumption is one of the reasoning tasks and it checks whether the subclass relation exists between two classes. We chose subsumption to evaluate the effectiveness of the proposed embeddings rather than link prediction because this task makes use of the normalized axioms to infer the subclass relation. The task of subsumption is reduced as a distance-based operation in the embedding vector space. Given a test instance of the form $C \sqsubseteq D$, we use D as source class and rank all other classes in the given ontology in an increasing order of their distance from D in the vector space. Based on the rank at which $C$ is present in the ranked list, we evaluate our model. We hypothesise that an embedding model that successfully captures the ontological information should be able to assign very close vector representations to the two classes in a subclass relation, hence, producing a lower rank for $C$.

In order to evaluate the performance of different models, we use Hits at ranks 1, 10 and 100 which report the fraction of the test cases where the given class C falls under top 1, 10 and 100 in rank list respectively. Median rank and 90$^{th}$ percentile rank were also considered to compare the overall performance of the models. A median rank of $m$ indicates that for 50\% of test cases, the correct answer was found below rank $m$. Similarly 90$^{th}$ percentile rank indicates the rank below which the correct class was found for 90\% of the test cases.

\begin{table}[ht]
\centering
\caption{Performance of Different Methods for Galen}
\label{table:galen-result}
\begin{tabular}{cccccc}
\hline
\textbf{Model}               & \textbf{Top1} & \textbf{Top10} & \textbf{Top100} & \textbf{Median} & \textbf{90th\% Rank} \\ \hline
TransE              & 0.00 & 0.00  & 0.00   & 10748       & 21308                \\ 
TransH              & 0.00 & 0.00  & 0.00   & 11721       & 21825                \\ 
DistMult            & 0.00 & 0.00  & 0.00   & 12600       & 21823                \\ 
EmEl                & 0.01 & 0.11  & 0.19   & 6039        & 21221                \\ 
EmEl++              & 0.02 & 0.11  & 0.16   & 6623        & 20635                \\ 
EmEl(var)           & \textbf{0.10} & \textbf{0.12}  & \textbf{0.26}   & \textbf{3540}        & \textbf{19669}                \\ \hline
\end{tabular}
\end{table}

\section{Results}
The evaluation results of the model on the subsumption task for all the three ontologies is given in Tables \ref{table:galen-result}, \ref{table:go-result} and \ref{table:snomed-result} respectively. In all the three cases, EmEl(var) outperforms almost every other model across all the five metrics. The variants of EmEl perform better than other traditional KG Embeddings i.e. TransE, TransH and DistMult showing the importance of capturing underlying structures of the ontologies. EmEl++, in general, tends to perform slightly better than EmEl on various metrics as well. EmEl(var) improves the performance on the Galen ontology compared to the previous best performing models. A notable improvement is seen in the Top1 and Top100 values i.e. Top1 score increases from 0.02 in EmEl++ to 0.10 in EmEl(var) which is quite significant. Similarly, we see an increase of Top100 score from 0.19 in EmEl to 0.26 in EmEl(var). The median rank becomes almost half of the previous best model and a significant improvement is also seen in 90th percentile rank.

\begin{table}[h!]
\centering
\caption{Performance of Different Methods for GO}
\label{table:go-result}
\begin{tabular}{cccccc}
\hline
\textbf{Model}               & \textbf{Top1} & \textbf{Top10} & \textbf{Top100} & \textbf{Median} & \textbf{90th\% Rank} \\ \hline
TransE              & 0.00 & 0.00  & 0.00   & 20079       & 40177                \\ 
TransH              & 0.00 & 0.00  & 0.00   & 26280       & 41996                \\ 
DistMult            & 0.00 & 0.00  & 0.00   & 22493       & 40425                \\ 
EmEl                & \textbf{0.01} & 0.08  & 0.15   & 9504        & 36447                \\ 
EmEl++              & \textbf{0.01} & \textbf{0.09}  & 0.15   & \textbf{7232}        & \textbf{33892}                \\ 
EmEl(var)           & \textbf{0.01} & \textbf{0.09}  & \textbf{0.17}   & 7542        & 34148                \\ \hline
\end{tabular}
\end{table}

Table \ref{table:go-result} shows a mixed result on the GO ontology. While EmEl(var) shows the best results for Top1, Top10 and Top100, it is comparable with EmEl++ in terms of the Median Rank and 90th Percentile Rank. 

\begin{table}[h!]
\centering
\caption{Performance of Different Methods for SNOMED CT}
\label{table:snomed-result}
\begin{tabular}{cccccc}
\hline
\textbf{Model}               & \textbf{Top1} & \textbf{Top10} & \textbf{Top100} & \textbf{Median} & \textbf{90th\% Rank} \\ \hline
TransE              & 0.00 & 0.00  & 0.00   & 150876      & 274465               \\ 
TransH              & 0.00 & 0.00  & 0.00   & 157186      & 278455               \\ 
DistMult            & 0.00 & 0.00  & 0.00   & 151624      & 275982               \\ 
EmEl                & 0.00 & 0.03  & 0.08   & 80289       & 277874               \\ 
EmEl++              & 0.00 & 0.03  & 0.06   & 87413       & 261359               \\ 
EmEl(var)           & \textbf{0.08} & \textbf{0.18}  & \textbf{0.36}   & \textbf{42759}       & \textbf{134829}                \\ \hline
\end{tabular}
\end{table}

Table \ref{table:snomed-result} contains results of our model on SNOMED CT, which is a large ontology. The Top1, Top10 and Top100 scores are significantly higher than any of the other models while the Median and 90th percentile ranks are less than half of the previous best performing models. This can be attributed to the fact that larger ontologies have larger relations that have many-to-many properties. EmEl(var) provides the model the freedom to incorporate those properties resulting in a substantially improved embedding. 

\section{Conclusion}
The existing Knowledge Graph and ontology embedding approaches assume that relations are one-to-one. This limits the possibility of using these embeddings for more expressive ontologies and for complex reasoning tasks. We have provided a simple yet effective method that overcomes this obstacle and helps embeddings to capture many-to-many relations. Through our evaluation, we have shown that our model outperforms the existing models in the subsumption task across three ontologies of varying sizes and characteristics. The technique that we described here, i.e., considering variance in relations, can be used in other knowledge graph embeddings such as TransE as well. The flexibility to model many-to-many relations also opens up the possibility of extending this work for more expressive description logics such as $SROIQ$.

\bibliographystyle{ACM-Reference-Format}
\bibliography{sample-base}


\begin{thebibliography}{22}


\ifx \showCODEN    \undefined \def \showCODEN     #1{\unskip}     \fi
\ifx \showDOI      \undefined \def \showDOI       #1{#1}\fi
\ifx \showISBNx    \undefined \def \showISBNx     #1{\unskip}     \fi
\ifx \showISBNxiii \undefined \def \showISBNxiii  #1{\unskip}     \fi
\ifx \showISSN     \undefined \def \showISSN      #1{\unskip}     \fi
\ifx \showLCCN     \undefined \def \showLCCN      #1{\unskip}     \fi
\ifx \shownote     \undefined \def \shownote      #1{#1}          \fi
\ifx \showarticletitle \undefined \def \showarticletitle #1{#1}   \fi
\ifx \showURL      \undefined \def \showURL       {\relax}        \fi
\providecommand\bibfield[2]{#2}
\providecommand\bibinfo[2]{#2}
\providecommand\natexlab[1]{#1}
\providecommand\showeprint[2][]{arXiv:#2}

\bibitem[\protect\citeauthoryear{Ali, Jabeen, Hoyt, and Lehman}{Ali
  et~al\mbox{.}}{2019}]%
        {ali2020keen}
\bibfield{author}{\bibinfo{person}{Mehdi Ali}, \bibinfo{person}{Hajira Jabeen},
  \bibinfo{person}{Charles~Tapley Hoyt}, {and} \bibinfo{person}{Jens Lehman}.}
  \bibinfo{year}{2019}\natexlab{}.
\newblock \showarticletitle{The KEEN Universe: An Ecosystem for Knowledge Graph
  Embeddings with a Focus on Reproducibility and Transferability}. In
  \bibinfo{booktitle}{\emph{International Semantic Web Conference, Springer}}.
\newblock


\bibitem[\protect\citeauthoryear{Baader, Brandt, and Lutz}{Baader
  et~al\mbox{.}}{2005}]%
        {baader}
\bibfield{author}{\bibinfo{person}{Franz Baader}, \bibinfo{person}{Sebastian
  Brandt}, {and} \bibinfo{person}{Carsten Lutz}.}
  \bibinfo{year}{2005}\natexlab{}.
\newblock \showarticletitle{Pushing the EL Envelope}. In
  \bibinfo{booktitle}{\emph{Proceedings of the 19th International Joint
  Conference on Artificial Intelligence}} (Edinburgh, Scotland)
  \emph{(\bibinfo{series}{IJCAI'05})}. \bibinfo{publisher}{Morgan Kaufmann
  Publishers Inc.}, \bibinfo{address}{San Francisco, CA, USA},
  \bibinfo{pages}{364–369}.
\newblock


\bibitem[\protect\citeauthoryear{Bordes, Usunier, Garcia-Duran, Weston, and
  Yakhnenko}{Bordes et~al\mbox{.}}{2013}]%
        {transe}
\bibfield{author}{\bibinfo{person}{Antoine Bordes}, \bibinfo{person}{Nicolas
  Usunier}, \bibinfo{person}{Alberto Garcia-Duran}, \bibinfo{person}{Jason
  Weston}, {and} \bibinfo{person}{Oksana Yakhnenko}.}
  \bibinfo{year}{2013}\natexlab{}.
\newblock \showarticletitle{Translating Embeddings for Modeling
  Multi-relational Data}. In \bibinfo{booktitle}{\emph{Advances in Neural
  Information Processing Systems}}, \bibfield{editor}{\bibinfo{person}{C.~J.~C.
  Burges}, \bibinfo{person}{L.~Bottou}, \bibinfo{person}{M.~Welling},
  \bibinfo{person}{Z.~Ghahramani}, {and} \bibinfo{person}{K.~Q. Weinberger}}
  (Eds.), Vol.~\bibinfo{volume}{26}. \bibinfo{publisher}{Curran Associates,
  Inc.}
\newblock
\urldef\tempurl%
\url{https://proceedings.neurips.cc/paper/2013/file/1cecc7a77928ca8133fa24680a88d2f9-Paper.pdf}
\showURL{%
\tempurl}


\bibitem[\protect\citeauthoryear{Consortium}{Consortium}{2004}]%
        {goc}
\bibfield{author}{\bibinfo{person}{Gene~Ontology Consortium}.}
  \bibinfo{year}{2004}\natexlab{}.
\newblock \showarticletitle{The Gene Ontology (GO) database and informatics
  resource}.
\newblock \bibinfo{journal}{\emph{Nucleic Acids Research}}
  \bibinfo{volume}{32} (\bibinfo{date}{01} \bibinfo{year}{2004}),
  \bibinfo{pages}{D258--D261}.
\newblock
\urldef\tempurl%
\url{https://doi.org/10.1093/nar/gkh036}
\showDOI{\tempurl}


\bibitem[\protect\citeauthoryear{Donnelly}{Donnelly}{2006}]%
        {snomed}
\bibfield{author}{\bibinfo{person}{Kevin Donnelly}.}
  \bibinfo{year}{2006}\natexlab{}.
\newblock \showarticletitle{SNOMED-CT: The advanced terminology and coding
  system for eHealth}.
\newblock \bibinfo{journal}{\emph{Studies in health technology and
  informatics}}  \bibinfo{volume}{121} (\bibinfo{date}{02}
  \bibinfo{year}{2006}), \bibinfo{pages}{279--90}.
\newblock


\bibitem[\protect\citeauthoryear{Ebrahimi, Eberhart, and Bianchi}{Ebrahimi
  et~al\mbox{.}}{2021}]%
        {Ebrahimi}
\bibfield{author}{\bibinfo{person}{Monireh Ebrahimi}, \bibinfo{person}{Aaron
  Eberhart}, {and} \bibinfo{person}{Federico Bianchi}.}
  \bibinfo{year}{2021}\natexlab{}.
\newblock \showarticletitle{Towards bridging the neuro-symbolic gap: deep
  deductive reasoners}.
\newblock \bibinfo{journal}{\emph{Applied Intelligence}}.
\newblock
\urldef\tempurl%
\url{https://doi.org/10.1007/s10489-020-02165-6}
\showDOI{\tempurl}


\bibitem[\protect\citeauthoryear{Grover and Leskovec}{Grover and
  Leskovec}{2016}]%
        {node2vec}
\bibfield{author}{\bibinfo{person}{Aditya Grover} {and} \bibinfo{person}{Jure
  Leskovec}.} \bibinfo{year}{2016}\natexlab{}.
\newblock \showarticletitle{Node2vec: Scalable Feature Learning for Networks}.
  In \bibinfo{booktitle}{\emph{Proceedings of the 22nd ACM SIGKDD International
  Conference on Knowledge Discovery and Data Mining}} (San Francisco,
  California, USA) \emph{(\bibinfo{series}{KDD '16})}.
  \bibinfo{publisher}{Association for Computing Machinery},
  \bibinfo{address}{New York, NY, USA}, \bibinfo{pages}{855–864}.
\newblock
\showISBNx{9781450342322}
\urldef\tempurl%
\url{https://doi.org/10.1145/2939672.2939754}
\showDOI{\tempurl}


\bibitem[\protect\citeauthoryear{Guarino, Oberle, and Staab}{Guarino
  et~al\mbox{.}}{2009}]%
        {Guarino}
\bibfield{author}{\bibinfo{person}{Nicola Guarino}, \bibinfo{person}{Daniel
  Oberle}, {and} \bibinfo{person}{Steffen Staab}.}
  \bibinfo{year}{2009}\natexlab{}.
\newblock \bibinfo{booktitle}{\emph{What Is an Ontology?}}
\newblock \bibinfo{pages}{1--17}.
\newblock
\urldef\tempurl%
\url{https://doi.org/10.1007/978-3-540-92673-3_0}
\showDOI{\tempurl}


\bibitem[\protect\citeauthoryear{Hogan, Blomqvist, Cochez, d'Amato, de~Melo,
  Gutierrez, Gayo, Kirrane, Neumaier, Polleres, Navigli, Ngomo, Rashid, Rula,
  Schmelzeisen, Sequeda, Staab, and Zimmermann}{Hogan et~al\mbox{.}}{2021}]%
        {hogan2021knowledge}
\bibfield{author}{\bibinfo{person}{Aidan Hogan}, \bibinfo{person}{Eva
  Blomqvist}, \bibinfo{person}{Michael Cochez}, \bibinfo{person}{Claudia
  d'Amato}, \bibinfo{person}{Gerard de Melo}, \bibinfo{person}{Claudio
  Gutierrez}, \bibinfo{person}{José Emilio~Labra Gayo},
  \bibinfo{person}{Sabrina Kirrane}, \bibinfo{person}{Sebastian Neumaier},
  \bibinfo{person}{Axel Polleres}, \bibinfo{person}{Roberto Navigli},
  \bibinfo{person}{Axel-Cyrille~Ngonga Ngomo}, \bibinfo{person}{Sabbir~M.
  Rashid}, \bibinfo{person}{Anisa Rula}, \bibinfo{person}{Lukas Schmelzeisen},
  \bibinfo{person}{Juan Sequeda}, \bibinfo{person}{Steffen Staab}, {and}
  \bibinfo{person}{Antoine Zimmermann}.} \bibinfo{year}{2021}\natexlab{}.
\newblock \bibinfo{title}{Knowledge Graphs}.
\newblock
\newblock
\showeprint[arxiv]{2003.02320}~[cs.AI]


\bibitem[\protect\citeauthoryear{Krötzsch, Simancik, and Horrocks}{Krötzsch
  et~al\mbox{.}}{2013}]%
        {description_primer}
\bibfield{author}{\bibinfo{person}{Markus Krötzsch},
  \bibinfo{person}{Frantisek Simancik}, {and} \bibinfo{person}{Ian Horrocks}.}
  \bibinfo{year}{2013}\natexlab{}.
\newblock \bibinfo{title}{A Description Logic Primer}.
\newblock
\newblock
\showeprint[arxiv]{1201.4089}~[cs.AI]


\bibitem[\protect\citeauthoryear{Kulmanov, Liu-Wei, Yan, and
  Hoehndorf}{Kulmanov et~al\mbox{.}}{2019}]%
        {Kulmanov}
\bibfield{author}{\bibinfo{person}{Maxat Kulmanov}, \bibinfo{person}{Wang
  Liu-Wei}, \bibinfo{person}{Yuan Yan}, {and} \bibinfo{person}{Robert
  Hoehndorf}.} \bibinfo{year}{2019}\natexlab{}.
\newblock \showarticletitle{EL Embeddings: Geometric Construction of Models for
  the Description Logic EL++}. In \bibinfo{booktitle}{\emph{Proceedings of the
  Twenty-Eighth International Joint Conference on Artificial Intelligence,
  {IJCAI-19}}}. \bibinfo{publisher}{International Joint Conferences on
  Artificial Intelligence Organization}, \bibinfo{pages}{6103--6109}.
\newblock
\urldef\tempurl%
\url{https://doi.org/10.24963/ijcai.2019/845}
\showDOI{\tempurl}


\bibitem[\protect\citeauthoryear{Lin, Liu, Sun, Liu, and Zhu}{Lin
  et~al\mbox{.}}{2015}]%
        {Lin2015}
\bibfield{author}{\bibinfo{person}{Yankai Lin}, \bibinfo{person}{Zhiyuan Liu},
  \bibinfo{person}{Maosong Sun}, \bibinfo{person}{Yang Liu}, {and}
  \bibinfo{person}{Xuan Zhu}.} \bibinfo{year}{2015}\natexlab{}.
\newblock \showarticletitle{Learning Entity and Relation Embeddings for
  Knowledge Graph Completion}. In \bibinfo{booktitle}{\emph{Proceedings of the
  Twenty-Ninth AAAI Conference on Artificial Intelligence}} (Austin, Texas)
  \emph{(\bibinfo{series}{AAAI'15})}. \bibinfo{publisher}{AAAI Press},
  \bibinfo{pages}{2181–2187}.
\newblock
\showISBNx{0262511290}


\bibitem[\protect\citeauthoryear{Lütfü~Özçep, Leemhuis, and
  Wolter}{Lütfü~Özçep et~al\mbox{.}}{2020}]%
        {ijcai2020-252}
\bibfield{author}{\bibinfo{person}{Özgür Lütfü~Özçep},
  \bibinfo{person}{Mena Leemhuis}, {and} \bibinfo{person}{Diedrich Wolter}.}
  \bibinfo{year}{2020}\natexlab{}.
\newblock \showarticletitle{Cone Semantics for Logics with Negation}. In
  \bibinfo{booktitle}{\emph{Proceedings of the Twenty-Ninth International Joint
  Conference on Artificial Intelligence, {IJCAI-20}}},
  \bibfield{editor}{\bibinfo{person}{Christian Bessiere}} (Ed.).
  \bibinfo{publisher}{International Joint Conferences on Artificial
  Intelligence Organization}, \bibinfo{pages}{1820--1826}.
\newblock
\urldef\tempurl%
\url{https://doi.org/10.24963/ijcai.2020/252}
\showDOI{\tempurl}
\newblock
\shownote{Main track.}


\bibitem[\protect\citeauthoryear{Mondal, Bhatia, and Mutharaju}{Mondal
  et~al\mbox{.}}{[n.d.]}]%
        {Mondal}
\bibfield{author}{\bibinfo{person}{Sutapa Mondal}, \bibinfo{person}{Sumit
  Bhatia}, {and} \bibinfo{person}{Raghava Mutharaju}.}
  \bibinfo{year}{[n.d.]}\natexlab{}.
\newblock \showarticletitle{EmEL++: Embeddings for EL++ Description Logic}. In
  \bibinfo{booktitle}{\emph{Proceedings of the AAAI 2021 Spring Symposium on
  Combining Machine Learning and Knowledge Engineering (AAAI-MAKE 2021)}},
  Vol.~\bibinfo{volume}{2846}.
\newblock


\bibitem[\protect\citeauthoryear{Nickel, Tresp, and Kriegel}{Nickel
  et~al\mbox{.}}{2011}]%
        {Nickel@icml}
\bibfield{author}{\bibinfo{person}{Maximilian Nickel}, \bibinfo{person}{Volker
  Tresp}, {and} \bibinfo{person}{Hans-Peter Kriegel}.}
  \bibinfo{year}{2011}\natexlab{}.
\newblock \showarticletitle{A Three-Way Model for Collective Learning on
  Multi-Relational Data}. In \bibinfo{booktitle}{\emph{Proceedings of the 28th
  International Conference on International Conference on Machine Learning}}
  (Bellevue, Washington, USA) \emph{(\bibinfo{series}{ICML'11})}.
  \bibinfo{publisher}{Omnipress}, \bibinfo{address}{Madison, WI, USA},
  \bibinfo{pages}{809–816}.
\newblock
\showISBNx{9781450306195}


\bibitem[\protect\citeauthoryear{Paszke, Gross, Massa, Lerer, Bradbury, Chanan,
  Killeen, Lin, Gimelshein, Antiga, Desmaison, Kopf, Yang, DeVito, Raison,
  Tejani, Chilamkurthy, Steiner, Fang, Bai, and Chintala}{Paszke
  et~al\mbox{.}}{2019}]%
        {pytorch}
\bibfield{author}{\bibinfo{person}{Adam Paszke}, \bibinfo{person}{Sam Gross},
  \bibinfo{person}{Francisco Massa}, \bibinfo{person}{Adam Lerer},
  \bibinfo{person}{James Bradbury}, \bibinfo{person}{Gregory Chanan},
  \bibinfo{person}{Trevor Killeen}, \bibinfo{person}{Zeming Lin},
  \bibinfo{person}{Natalia Gimelshein}, \bibinfo{person}{Luca Antiga},
  \bibinfo{person}{Alban Desmaison}, \bibinfo{person}{Andreas Kopf},
  \bibinfo{person}{Edward Yang}, \bibinfo{person}{Zachary DeVito},
  \bibinfo{person}{Martin Raison}, \bibinfo{person}{Alykhan Tejani},
  \bibinfo{person}{Sasank Chilamkurthy}, \bibinfo{person}{Benoit Steiner},
  \bibinfo{person}{Lu Fang}, \bibinfo{person}{Junjie Bai}, {and}
  \bibinfo{person}{Soumith Chintala}.} \bibinfo{year}{2019}\natexlab{}.
\newblock \showarticletitle{PyTorch: An Imperative Style, High-Performance Deep
  Learning Library}.
\newblock In \bibinfo{booktitle}{\emph{Advances in Neural Information
  Processing Systems 32}}, \bibfield{editor}{\bibinfo{person}{H.~Wallach},
  \bibinfo{person}{H.~Larochelle}, \bibinfo{person}{A.~Beygelzimer},
  \bibinfo{person}{F.~d\textquotesingle Alch\'{e}-Buc},
  \bibinfo{person}{E.~Fox}, {and} \bibinfo{person}{R.~Garnett}} (Eds.).
  \bibinfo{publisher}{Curran Associates, Inc.}, \bibinfo{pages}{8024--8035}.
\newblock
\urldef\tempurl%
\url{http://papers.neurips.cc/paper/9015-pytorch-an-imperative-style-high-performance-deep-learning-library.pdf}
\showURL{%
\tempurl}


\bibitem[\protect\citeauthoryear{Rector, Rogers, and Pole}{Rector
  et~al\mbox{.}}{1996}]%
        {galen}
\bibfield{author}{\bibinfo{person}{Al Rector}, \bibinfo{person}{Je Rogers},
  {and} \bibinfo{person}{P Pole}.} \bibinfo{year}{1996}\natexlab{}.
\newblock \bibinfo{title}{The GALEN high level ontology}.
\newblock
\newblock


\bibitem[\protect\citeauthoryear{Singh, Mondal, Bhatia, and Mutharaju}{Singh
  et~al\mbox{.}}{2021}]%
        {Singh_Mondal_Bhatia_Mutharaju_2021}
\bibfield{author}{\bibinfo{person}{Gunjan Singh}, \bibinfo{person}{Sutapa
  Mondal}, \bibinfo{person}{Sumit Bhatia}, {and} \bibinfo{person}{Raghava
  Mutharaju}.} \bibinfo{year}{2021}\natexlab{}.
\newblock \showarticletitle{Neuro-Symbolic Techniques for Description Logic
  Reasoning (Student Abstract)}.
\newblock \bibinfo{journal}{\emph{Proceedings of the AAAI Conference on
  Artificial Intelligence}} \bibinfo{volume}{35}, \bibinfo{number}{18}
  (\bibinfo{date}{May} \bibinfo{year}{2021}), \bibinfo{pages}{15891--15892}.
\newblock
\urldef\tempurl%
\url{https://ojs.aaai.org/index.php/AAAI/article/view/17942}
\showURL{%
\tempurl}


\bibitem[\protect\citeauthoryear{Smaili, Gao, and Hoehndorf}{Smaili
  et~al\mbox{.}}{2018}]%
        {ontovec}
\bibfield{author}{\bibinfo{person}{Fatima~Zohra Smaili}, \bibinfo{person}{Xin
  Gao}, {and} \bibinfo{person}{Robert Hoehndorf}.}
  \bibinfo{year}{2018}\natexlab{}.
\newblock \showarticletitle{{Onto2Vec: joint vector-based representation of
  biological entities and their ontology-based annotations}}.
\newblock \bibinfo{journal}{\emph{Bioinformatics}} \bibinfo{volume}{34},
  \bibinfo{number}{13} (\bibinfo{date}{06} \bibinfo{year}{2018}),
  \bibinfo{pages}{i52--i60}.
\newblock
\showISSN{1367-4803}
\urldef\tempurl%
\url{https://doi.org/10.1093/bioinformatics/bty259}
\showDOI{\tempurl}
\showeprint{https://academic.oup.com/bioinformatics/article-pdf/34/13/i52/25098469/bty259\_hoehndorf.30.sup.1.pdf}


\bibitem[\protect\citeauthoryear{Trouillon, Welbl, Riedel, Gaussier, and
  Bouchard}{Trouillon et~al\mbox{.}}{2016}]%
        {pmlr-v48-trouillon16}
\bibfield{author}{\bibinfo{person}{Théo Trouillon}, \bibinfo{person}{Johannes
  Welbl}, \bibinfo{person}{Sebastian Riedel}, \bibinfo{person}{Eric Gaussier},
  {and} \bibinfo{person}{Guillaume Bouchard}.} \bibinfo{year}{2016}\natexlab{}.
\newblock \showarticletitle{Complex Embeddings for Simple Link Prediction}. In
  \bibinfo{booktitle}{\emph{Proceedings of The 33rd International Conference on
  Machine Learning}} \emph{(\bibinfo{series}{Proceedings of Machine Learning
  Research}, Vol.~\bibinfo{volume}{48})},
  \bibfield{editor}{\bibinfo{person}{Maria~Florina Balcan} {and}
  \bibinfo{person}{Kilian~Q. Weinberger}} (Eds.). \bibinfo{publisher}{PMLR},
  \bibinfo{address}{New York, New York, USA}, \bibinfo{pages}{2071--2080}.
\newblock
\urldef\tempurl%
\url{http://proceedings.mlr.press/v48/trouillon16.html}
\showURL{%
\tempurl}


\bibitem[\protect\citeauthoryear{Wang, Zhang, Feng, and Chen}{Wang
  et~al\mbox{.}}{2014}]%
        {transh}
\bibfield{author}{\bibinfo{person}{Zhen Wang}, \bibinfo{person}{J. Zhang},
  \bibinfo{person}{Jianlin Feng}, {and} \bibinfo{person}{Z. Chen}.}
  \bibinfo{year}{2014}\natexlab{}.
\newblock \showarticletitle{Knowledge Graph Embedding by Translating on
  Hyperplanes}. In \bibinfo{booktitle}{\emph{AAAI}}.
\newblock


\bibitem[\protect\citeauthoryear{Yang, tau Yih, He, Gao, and Deng}{Yang
  et~al\mbox{.}}{2015}]%
        {yang2015embedding}
\bibfield{author}{\bibinfo{person}{Bishan Yang}, \bibinfo{person}{Wen tau Yih},
  \bibinfo{person}{Xiaodong He}, \bibinfo{person}{Jianfeng Gao}, {and}
  \bibinfo{person}{Li Deng}.} \bibinfo{year}{2015}\natexlab{}.
\newblock \bibinfo{title}{Embedding Entities and Relations for Learning and
  Inference in Knowledge Bases}.
\newblock
\newblock
\showeprint[arxiv]{1412.6575}~[cs.CL]


\end{thebibliography}

\appendix









\end{document}